# Safety Prioritized, Reinforcement Learning Enabled, Traffic Flow Optimization In 3D City-Wide Simulation Environment

Mira Nuthakki *

*Corresponding author

miranuthakki@gmail.com

## Abstract

Traffic congestion and collisions represent significant economic, environmental, and social challenges worldwide. Traditional traffic management approaches have shown limited success in addressing these complex, dynamic problems. To address the current research gaps, three potential tools are developed: a comprehensive 3D city-wide simulation environment that integrates both macroscopic and microscopic traffic dynamics; a collision model; and a reinforcement learning framework with custom reward functions prioritizing safety over efficiency. Unity game engine-based simulation is used for direct collision modeling. A custom reward enabled reinforcement learning method, proximal policy optimization (PPO) model, yields substantial improvements over baseline results, reducing the number of serious collisions, number of vehicle-vehicle collisions, and total distance travelled by over 3 times the baseline values. The model also improves fuel efficiency by 39% and reduces carbon emissions by 88%. Results establish feasibility for city-wide 3D traffic simulation applications incorporating the vision-zero safety principles of the Department of Transportation, including physics informed, adaptable, realistic collision modeling, as well as appropriate reward modeling for real-world traffic signal light control towards reducing collisions, optimizing traffic flow and reducing greenhouse emissions.

<u>Keywords</u>: reinforcement learning, traffic simulation, vehicle safety, traffic flow optimization, deep learning, intelligent transportation systems

## Introduction

Traffic congestion and crashes are escalating global challenges, with approximately 286 million vehicles operating on U.S. roads in 2023 and 1.5 billion globally – a substantial increase from 193 million and 500 million, respectively, in 1990[1,2,3]. This is likely related to population growth, urban and suburban migration for jobs, older traffic infrastructure and existing suboptimal traffic management. These conditions generate enormous economic burdens: $87 billion in lost productivity in the United States[4] and over €110 billion annually in Europe[5], while simultaneously exacerbating environmental impacts via contributions to urban heat islands and the global 2.4°C temperature increase between 1847 and 2013[6].

The human and economic toll is equally concerning. Traffic crashes result in $277 billion in property damage and an additional $594 billion attributed to loss of life and decreased quality of life[7]. In 2020 alone, over 5 million collisions occurred in the U.S., including 35,000 fatalities[8]. Fatal accidents increased 7% from 2019-2020 and 10% from 2020-2021, with approximately 19,000 crashes occurring daily in the United States and 1 fatal accident every 12 minutes[9].

Globally, traffic crashes cause approximately 1.19 million deaths annually, with 20-50 million people suffering injuries. These incidents disproportionately affect vulnerable populations; notably, 90% of fatalities occur in low/middle-income countries. Road traffic injuries are the leading cause of death for those aged 5-29, and two-thirds of fatalities affect working-age individuals (18-59)[10]. With global urban populations projected to increase from 55% to 68% by 2050[11], the need for intelligent transportation management systems has become critical.

Public transportation and electric shared transportation can help mitigate traffic issues but are not accessible to many people, due to infrastructure costs and/or lack of community/political will. Traditional statistical methods of traffic flow modeling include Webster's cycle length optimization[12], VA (Vehicle-Actuated) signal control[13], ALINEA ("Asservissement Linéaire d'Entrée Autoroutière" translating to "Linear Control of Highway Entrance")[14], adaptive systems such as SCATS (Sydney Coordinated Adaptive traffic System)[15] and self-organizing traffic lights (SOTLs, which give preference to large groups of vehicles that have been waiting the longest)[16], SPECIALIST (SPEed (limit) ControllIng ALgorIthm using Shock wave Theory)[17], and CityFlow for 2D traffic simulation[18]. These approaches have been hampered by limited use and inability to adapt to dynamic demands or be dynamically updated; meanwhile, congestion continues to worsen.

Deep learning methods in research, particularly CNN (Convolutional Neural network) and LSTM (Long Short-Term Memory) architectures, have demonstrated superior performance over traditional statistical approaches[19,20], with recursive algorithms generally outperforming convolutional methods[21]. Real-world implementations have been limited to some Asian cities such as Alibaba's CityBrain (a CNN- autoencoder model) in Hangzhou, that have improved the city's standings from 5th to 57th in congestion rankings, with an approximately 15% reduction in congestion[22]. This impactful though not exceptional performance may be related to the difficulty of optimizing traffic flows given the nuanced relationships between road sections and traffic patterns[23], failure to account for nonlinear and complex traffic dynamics[24,25], inadequate consideration of unpredictable non labelled behavior[26], and prohibitive computational costs[27,28].

In addressing some of the above issues, Reinforcement learning (RL) offers significant advantages to other deep learning methods. It reduces online computation requirements[29]. RL has been successfully applied to simulation based traffic signal control, ramp metering, variable speed limit establishment, and vehicle motion control[30,31,32,33], with recent enhancements to performance, convergence rates, and scalability[34,35,36,37]. In RL implementations, an agent controls the traffic environment, making decisions based on current states to maximize rewards over time. RL methods are not reliant on human-labor intensive labelled data needed by other deep learning architectures. Unlike large language models, which are limited by slow inference times due to high computational complexity[38], RL models have faster inference times (essential in real-time traffic signal control) and support human control in the form of custom rewards in dynamic high-stakes scenarios.

RL models typically require initial offline training with simulated data before transfer training in real online environments. This allows training with large amounts of variable data and over long periods. This approach is valuable because direct training on limited data in real-world systems could create hazardous conditions. This methodology requires effective simulation environments that, despite being stochastic, provide sufficient training opportunities in conjunction with congestion, safety, and capacity optimization[29].

The current traffic simulation approaches fall into two categories: macroscopic models (e.g., CTM (Cell Transmission Model), MFD (macroscopic Fundamental Diagram), METANET (Macroscopic Engine for Traffic Analysis and Network Evaluation Tool)) that simulate aggregate traffic flows; and microscopic models (e.g., SUMO (Simulation of Urban MObility), MATSim (Multi-Agent Transport SIMulation), VISSIM (Verkehr In Städten - SIMulationsmodell), AIMSUN (Advanced Interactive Microscopic Simulator for Urban and Non-Urban Networks)) that simulate individual vehicles and their interactions. Macroscopic models offer tractability and interpretability but work under assumptions of traffic homogeneity. Microscopic simulations provide detailed intersection-level behavioral insights but lack consideration of big-picture or influence of upstream/downstream parameters[39]. Microscopic models fragment vehicle control across multiple sub-models, while macroscopic approaches oversimplify network dynamics. Moreover, both lack flexibility regarding driver heterogeneity[40], and neither models traffic in three dimensions. Notably, a 3D approach would enhance visual interpretability, transparency and accountability for operators, citizens and decision-makers, including those without technical knowledge.

Game engines like Unity offer promising alternatives for creating realistic simulation models, supporting both 2D and 3D modeling while enabling physics-informed simulations with multiple vehicle parameters[41]. To date, however, only a few studies have explored the use of Unity for traffic signal control, and were focused mostly on single intersections or roads[42,43].

The U.S. Department of Transportation (DOT) emphasizes Vision Zero principles which call for zero traffic-related deaths through safety-prioritized traffic flow management and infrastructure changes. This is a realistic goal, as demonstrated by Hoboken, New jersey accomplishing 0 traffic-related deaths since 2017 through the utilization of Vision Zero principles. Yet, both traditional micro- and macro simulation environments are programmed to avoid vehicle collisions; collision modeling is not possible; some chosen vehicles can be programmed to ignore other vehicles at an intersection, but this is not realistic or scalable. This has unfortunately limited collision-focused research. Surrogate safety metrics such as time to collision (TTC), post-encroachment time (PET), and deceleration required to avoid a crash (DRAC) are calculated instead[44]. These metrics are inherently limited as their underlying parameters are optimized for collision avoidance and may reflect "noise"/random variations in traffic simulation. A dedicated physics-informed game engine would allow consideration of different types of realistic collisions and aid in the incorporation of safety into traffic control studies.

In this feasibility study, three key objectives are explored: 1) development of a micro- and macro-3D simulation environment using a game engine to simulate a small to moderate-sized city traffic network; this would address drawbacks associated with fragmented vehicle control under various sub-models seen in microscopic environments and lack of consideration of heterogeneity in overall traffic flows seen in macroscopic environments, by formulation of a single simulation model that controls and is affected by all vehicle and environment parameters while taking into account overall traffic flows and heterogeneity, with the 3D aspect allowing for intuitive interpretability for technologists and non-technologists; 2) real–life collision modeling and mitigation, a capability lacking in existing simulation environments despite the need to incorporate Vision Zero principles; and 3) development of a custom reinforcement learning model for traffic signal light control in a sample physics-informed collision-heavy 3D large urban environment that would improve both collisions and traffic congestion.

**METHODS**

A simulation city environment was created, incorporating a baseline City Generator asset in Unity 2022.3. The city generator allows for creation of various city sizes based on distance to center parameter. Residential, commercial, corner, block-spanning buildings were incorporated, with higher residential density occurring at greater distances from the city center (Supplement section 1).

**Road network, connectivity and intersection logic:** Each road segment was implemented as a waypoints container object, containing a sequential list of waypoint transforms. Each waypoint maintains three pieces of information: vector position in world space; rotation representing the forward direction of travel (as a 4 element vector, quaternion); and a reference to its parent container. Waypoints were created and positioned during the environment construction phase, with additional waypoints manually generated through the editor interface by holding the Shift key and clicking on the terrain.

During initialization, each waypoint in a container was oriented to face the next point in sequence. Each waypoint container maintains an array of potential continuation paths accessible from its final waypoint (a "next-way" array containing potential next ways), calculated as follows: the first waypoints of each next available path in the intersection scene are collected. For each such candidate waypoint, the euclidean distance from the current final waypoint is calculated. Candidate waypoints are filtered using a proximity threshold between 8 and 35 units to identify potential connected paths. This range ensured that only relevant nearby paths were considered while preventing connections to distant road segments. For each proximate waypoint, the angular relationship was calculated: for typical right-sided driving traffic, accepted angles were between 340-360° or 0-80°. The next-way array is then populated with references to valid continuation path containers, enabling vehicles to make realistic turning decisions at intersections. The waypoint system utilized a Gizmo-based visualization system provided by Unity for editor-time debugging.

**Vehicle physics** implementation utilized Unity's built-in wheel transforms (for front-right (FR), front-left (FL), back-right (BR), and back-left (BL) wheel). Each vehicle utilized corresponding four WheelCollider components configured with calibrated parameters:

- Suspension Spring Strength: The baseline value of 25,000 was used for most vehicles, with the configurable range of 10,000-60,000 available to create vehicles with varying suspension stiffness. Higher values created stiffer suspensions appropriate for heavier vehicles, while lower values created softer suspensions suitable for lighter vehicles.
- Suspension Dampers: The baseline value of 1,500 with a range of 1,000-6,000 controlled the damping rate of the suspension. Higher values reduced oscillation but increased rigidity, while lower values allowed more oscillation but improved terrain adaptation.
- Suspension Distance: Fixed at 0.05 units to create a consistent ground clearance across all vehicle types.
- Wheel Radius: Dynamically calculated from the mesh bounds of the wheel model multiplied by the local (city-size) scale
- Wheel Mass: Set to 1,500 units for all wheels to ensure consistent physic behavior. • Motor Torque: Applied only to front wheels (indices 0 and 1 in the wheel array) and calculated as an inverse linear interpolation based on current speed. This realistic power curve decreased as the vehicle approached its speed limit. The base carPower parameter (120) was multiplied by 30 to convert to an appropriate torque range for the physics system.
- Braking Torque: Applied to all wheels when braking was necessary, with intensity determined by obstacle proximity or speed limit enforcement. The base brakePower (8) was multiplied by the brake factor (range: 0-6000, derived from obstacle raycasts). This created a proportional braking response that intensified as obstacles approached.
- Vehicle stability was enhanced by adjusting the center of mass to a slightly lower position (-0.05), preventing roll-over during cornering
- Steering: The maxSteerAngle parameter (35-72°) was dynamically calculated based on vehicle wheelbase to ensure appropriate turning characteristics. The target waypoint position was transformed into the vehicle's local coordinate space, creating a vector where the x component represented the lateral offset to the target, and the z component represented the forward distance to the target. Steering angle was derived from the normalized x component scaled by the maximum steering angle. This produced a proportional steering response where targets directly ahead resulted in zero steering; targets to the right produced positive steering angles; targets to the left produced negative steering angles. More extreme lateral positions produced stronger steering responses. The calculated steering angle was applied to the physical steering angle of the front wheel colliders, the orientation of the forward obstacle raycast origin and the visual rotation of the steering wheel model.

**Vehicle navigation** utilized the target-based steering system updated at fixed intervals (0.02 seconds). The current target was determined by the waypoints array of the vehicle's assigned path container. When the vehicle came within 5 units of the current waypoint, progression (moving forward) occurred. Upon reaching the final waypoint in a path, the vehicle selected a new path from the available next-way options using random selection to distribute traffic naturally.

**Vehicle speed** was regulated through a combination of motor torque and braking. Forward movement was achieved by applying motor torque to the front wheels. Braking was triggered by either detection of obstacles through raycasts, or exceeding the vehicle speed limit. When braking was required, motor torque was set to zero and brake torque was applied.

**Obstacle detection and avoidance** utilized a three-directional raycast system implemented on a 4-frame cycle. The central raycast oriented from the front of the vehicle along the current steering angle, extending 6 units ahead. 2 side raycasts were angled at ±37° (tangent 0.75) from the forward direction, extending 2 units. The maximum brake value from all three raycasts was applied. This 3-directional detection system created realistic obstacle avoidance behavior while maintaining acceptable computational overhead.

**Collision Physics:** In Unity's physics engine (PhysX), vehicle collisions occur when the collider components of two or more GameObjects intersect in 3D space. Each vehicle was equipped with a primary rigidbody collider representing the vehicle's chassis and four WheelCollider components representing the wheels. Collisions occurred for several physics-based reasons:

- Momentum and Speed: Vehicles traveling at higher speeds had greater momentum, making them less responsive to steering and braking inputs. This often led to collisions in situations requiring sudden deceleration, particularly when approaching a stopped vehicle, encountering sharp turns or reacting to sudden traffic light changes, similar to real life.
- Limited Perception Distance: The raycast system had a maximum detection range of 6 units for forward detection and 2 units for side detection. When vehicles were traveling faster than could be safely stopped within this perception distance, collisions became unavoidable. This mirrors real-world limitations in driver perception and reaction time.
- Waypoint Following Limitations: Vehicles steering toward waypoints sometimes prioritized reaching the waypoint over collision avoidance. When multiple vehicles converged on the same waypoint at an intersection, the path-following algorithm could produce competing trajectories that resulted in collisions.
- Physics Solver Limitations: Unity's fixed timestep physics solver (typically running at 0.02-second intervals) sometimes struggled to resolve complex multi-vehicle interactions, particularly in congested areas. This occasionally resulted in penetration between colliders before the physics system could apply appropriate separation forces.
- Weight and Inertia: Vehicle mass (set to 4000 units) and the lowered center of mass created significant inertia that affected stopping distance and turning capability. While this improved stability, it also made vehicles less maneuverable in emergency situations.

**Collision response physics:** When collisions occurred, Unity's physics engine handled the initial response by calculating collision forces based on the relative mass, velocity, and angle of impact; and applying impulse forces to the rigidbodies involved. It triggered collision callbacks that the simulation used to track and respond to collision events. The code then handled these physics events through three primary callbacks:

- OnCollisionEnter: Detected initial contact and categorized collisions as vehicle to vehicle (when the other object had the "vehicle" tag), or vehicle to non-vehicle collisions (when the other object had all other tags)
- OnCollisionStay: Monitored ongoing collisions and assessed their severity by tracking stopped time during collision. Side raycasts were disabled after 7 seconds of continuous stopping to allow vehicles more freedom to maneuver in tight spaces. When a vehicle remained stopped for >30 seconds, they were labeled as serious collisions, and if they remained stopped for >60 seconds, the vehicles were removed from the environment, simulating real life towing after a major collision. These mechanisms prevented permanent gridlock while still allowing for natural congestion and resolution patterns to emerge in the simulation.
- OnCollisionExit: Reset all collision flags/labels.

This combination of Unity's physics-based collision detection and our custom response handlers created a realistic traffic environment where collisions occurred naturally from physical interactions and traffic congestion, rather than being artificially scripted.

**Traffic Density Control:** Vehicles were instantiated on waypoint paths by locating all waypoints container objects in the scene, representing available vehicle paths. For each path, a randomly selected vehicle model was instantiated at the first waypoint. If intense traffic parameter was enabled, additional vehicles were placed at 40% of the distance between the first two waypoints when that distance exceeded 50 units. The instantiation process could be repeated through multiple iterations (50 cycles by default). The system also allowed for manually duplicating vehicles and their waypoint containers ahead or behind each other.

**Traffic Lights:** Green signal phase duration is the primary method of traffic signal control with a range of 5-60 seconds. Transition phases consisted of a 3 second yellow phase. This was followed by all red interval of 1 second, before the perpendicular traffic light at the intersection initiated it's green phase.

**Metrics:** Data was recorded to CSV files at configurable intervals for post-simulation analysis. The testing configuration used a 600-second simulation duration with metrics capture at the conclusion, while the training configuration used continuous metrics capture at configurable intervals during extended simulations. These included:

- Speed distribution across defined ranges (percentage of time in each speed bin of 0-5, 5-10, 10-15, 15-20, 20-25, 25-30, 30-35 units/sec)
- Total distance traveled per vehicle
- Total time stopped (consecutively and cumulatively)
- Collision counts (vehicle-vehicle, vehicle-non-vehicle and total collisionsl) •

Pass-through counts at traffic signals
- Duration in each traffic light signal state

**Reinforcement Learning (RL):** RL takes the information (observations) from the environment and decides on an action on the environment via it's policy (process for deciding actions). It then observes the new resultant state and determines the next action (Supplement Fig 2). Observations collected were the position vectors of vehicles and traffic lights, as well as the current signal phase and green light duration. These were processed by a CNN architecture. Actions (guided by rewards) modified traffic green light signal duration and vehicle speed limits. Episode ran for 600 seconds as default. Each episode was reset with random green light durations 5 to 60 seconds, and speed limit (20-35 units/sec, simulating 20-35 mph). To enable realistic traffic light cycles while maintaining computational efficiency, green light duration decisions were triggered at 60-second intervals; light phase cycling was not changed.

A few RL methods were evaluated: 1. Proximal Policy Optimization (PPO): An on-policy algorithm that learns value functions from current policy observations, featuring entropy regularization to encourage exploration (avoiding local minima of the loss function by choosing actions with higher entropy/ randomness, so that a global minimum can be achieved). On-policy algorithms tend to be more stable but data-hungry. 2. Soft Actor-Critic (SAC): An off-policy algorithm using observations from previous policy explorations, with entropy maximization (maximizing randomness to get the maximum reward) 3. Multi-Agent Posthumus Credit Assignment (MA-POCA): A centralized critic approach that provides team-wide rewards, enabling agents to learn the value of group-beneficial actions even if they result in individual sacrifices. Curriculum learning was also attempted through increasing vehicles.

**Reward Structure:** The reward system combined continuous and discrete elements: Continuous rewards included time stopped* $-10^{-5}$, distance traveled* $+10^{-8}$, and speed 25-30 timer* $+10^{-5}$. Discrete rewards included +0.01 for passing through an intersection, -1 for a serious collision, -0.01 for all vehicle collisions.

**Losses and hyperparameters:** Losses monitored were policy loss (how much the policy is changing) and value loss (how well the model is able to predict the value of each state). The hyperparameter configuration (complete list in supplement section 6) included buffer_size (how many experiences (agent observations, actions and rewards obtained) to be collected before any learning or updating policy): 102400; learning_rate (how much the policy changes): 0.0003; beta: 0.05 (to encourage entropy); epsilon (policy change cap rate): 0.2; lambda (how much the agent relies on the current value estimate/high bias vs actual rewards/ high variance): 0.95.

## RESULTS

**Simulation Environment:** The main simulation environment for this study consisted of a small to moderate-sized urban city area with 46 traffic lights (in X and T configurations) and 536-879 vehicles (Fig 1a,b). The road network consisted of approximately 136 distinct waypoint containers, each containing between 2-12 waypoints depending on the length and complexity of the road segment. Configurable traffic density and random paths allowed for vehicular and traffic flow heterogeneity, taking into account both macro and micro factors. Subsequent RL models were able to learn from all types of scenarios such as regular traffic, peak congestion, free flows (such as at night time) and local collision induced bottlenecks that had a ripple effect on overall traffic. The current collision heavy simulation environment ran for 10min, at 20 times speed (representing 3.3 hours of real traffic time).

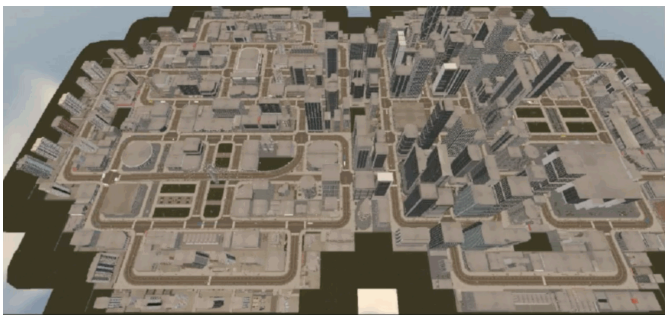
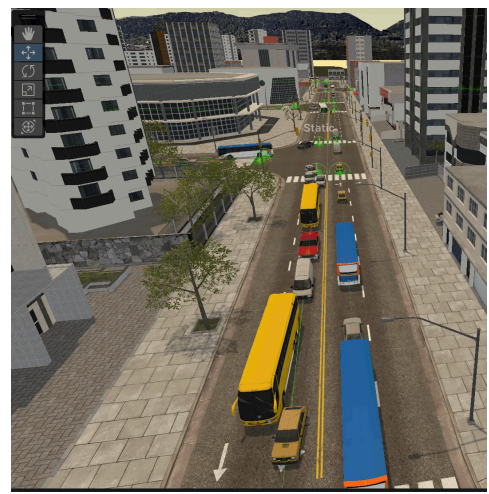

Fig 1a (top): Overview of downtown and surrounding areas of a small city.
Fig 1b (right): A street lined with school buses and office traffic during peak congestion

**Collision Modeling**: Physics enabled collisions were characterized as minor (such as representing fender-benders, minor damage to frame/suspension) that were able to be extricated out of in under 30 seconds by the vehicles themselves (representing stopping/ marked slowing by both the involved vehicles and the surrounding vehicles to enable the involved vehicles to clear out of the way of incoming traffic and rejoin regular traffic). They would be characterized as serious when they lasted beyond 30 seconds. These stayed in position for an additional 30 seconds (total 60 seconds) before the environment removed them, simulating a major collision which causes progressive downstream congestion as the vehicles are too damaged to extricate themselves without towing help that takes some time to arrive and clear.

Given the inability to determine statistical significance with realistically rare collision events by tight control of vehicle dynamics (Supplement Fig 15), a high collision environment was allowed to test collision reduction within time constraints. Collisions varied with each simulation as they were physics enabled responding to random vehicular paths but for the simulation environment configured for this study, averaged around 400 for serious collisions, 5000 for total vehicle vehicle collisions and 1500 for total vehicle to non-vehicle collisions. Similar to real life, majority of the collisions, especially minor collisions, were rear-end collisions happening while approaching turns, and resolved within 10-15 seconds (~80%). Sideswipe (side on side) collisions were less likely to result in serious collisions than T-bone (side on head) collisions. Head on head collisions were rare. Approximately 80% of serious collisions were vehicle-vehicle interactions, and over 90% of vehicle to non-vehicle collisions (such as curb riding) were minor.

**Model Performance:** In the initial model selection phase with 536 vehicles, PPO demonstrated earlier, higher, and more consistent cumulative rewards with higher value estimates and moderate value loss compared to the SAC and MA-PPO models (Supplement Fig 3). The selected model, centralized PPO (henceforth referenced as the model), was then trained with data for 879 vehicles. During training, the model maintained stable policy loss below a value of 1, demonstrated appropriate increasing and then plateauing cumulative and curiosity rewards, and showed gradually decreasing value loss (Supplement Fig. 4).

In testing (averaged over 3 random seed trials), the model increased total distance travelled by all cars and average distance travelled by each car by 345%, from 338.62 to 1509.23 unity units/meters (Fig. 2). The model reduced serious collisions by 75%, from 424 to 106 (Fig 3). The total number of vehicle-vehicle collisions decreased by 79%, from 5082 to 1046 (Fig. 4). The total number of vehicle to non-vehicle collisions decreased most, by 96%, from 1576 to 60 collisions (Fig. 5).Time spent across all speed ranges increased after training. Specifically, time per vehicle in the speed ranges 5-10, 10-15, 15-20, 20-25, 25-30, and 30-35 units/sec increased by 1.83, 3, 3.62, 5.6, 13.25, and 3.5 times respectively, with most improvements in the 20-30 units/sec range (Fig. 6).

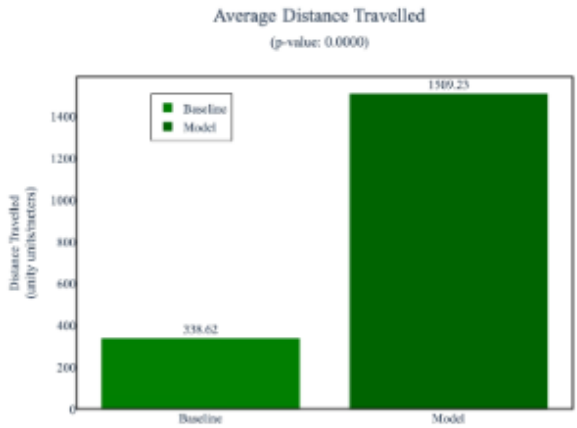
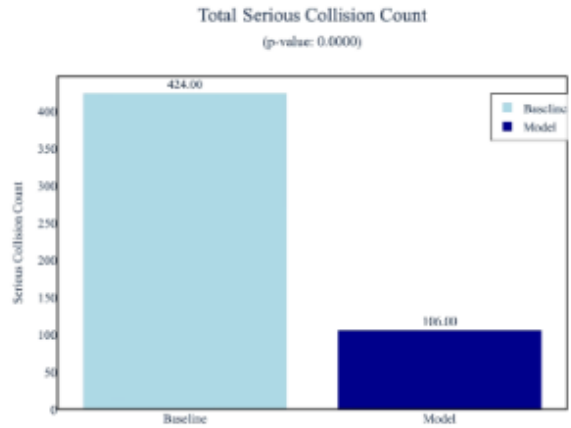

Fig 2: Average distance (meters/unity units) travelled in baseline vs model. Improvement by model was 345%, from 338.62m to 1509.23m

Fig 3. Total serious collisions count in baseline vs model. Improvement by model was 75%, from 424 to 106

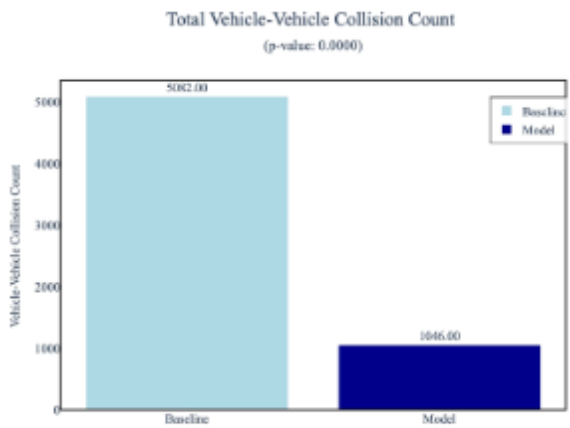
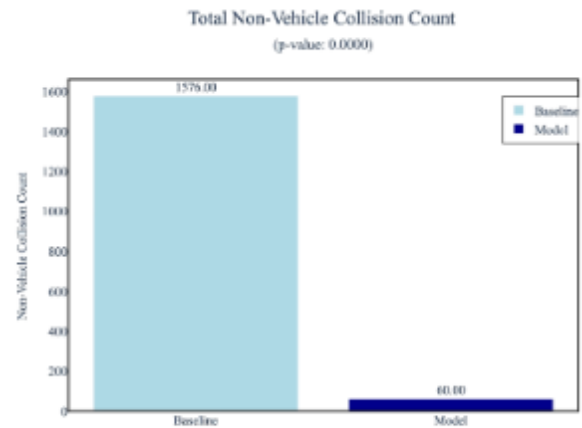

Fig 4. Total non-serious vehicle-vehicle collision in baseline vs model. Improvement by model was 79%, from 5082 to 1046

Fig 5. Vehicle to non-vehicle collisions in baseline vs model. Improvement by model was 96%, from 1576 to 60 minor collisions.

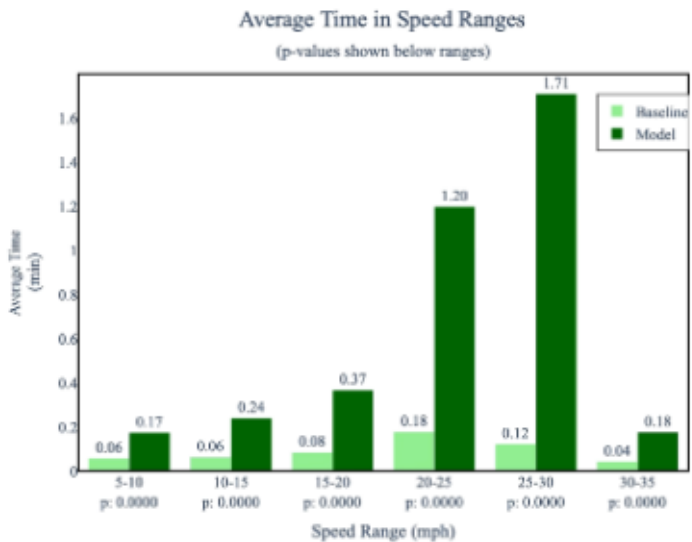

Fig 6. Average time in min over speed ranges 0-5 mph, 5-10 mph, 10-15mph, 15-20mph, 20-25mph, 25-30mph, 30-35mph in baseline vs model. Model improved vehicular speed, 200-1300%, most in the 25-30mph range.

While total stopped time increased by 110%, from 2.18 min in the baseline to 4.59 min in the model results, this counterintuitive result is related to vehicle survival rates (Figs. 7 and 8a, b). In the baseline environment, many vehicles are removed early due to their involvement in serious collisions, resulting in fewer remaining vehicles and consequently less congestion in later periods. In the model-controlled environment, there were less vehicles removed from less serious collisions, and more active vehicles remained active throughout the testing period, including at the end of the testing period.

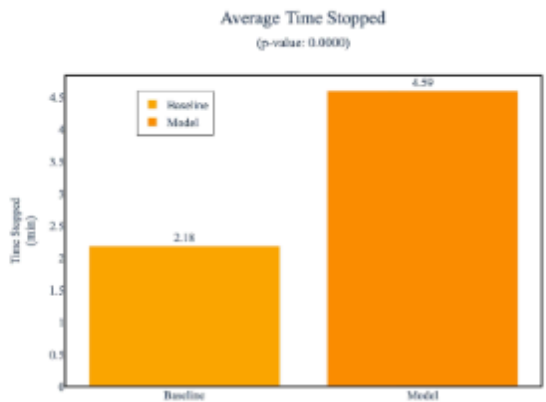

Fig 7. Average time (min) stopped per car in baseline vs model. There was an increase of 110%, from 2.18min in baseline to 4.59min with the model, explained by Fig 7

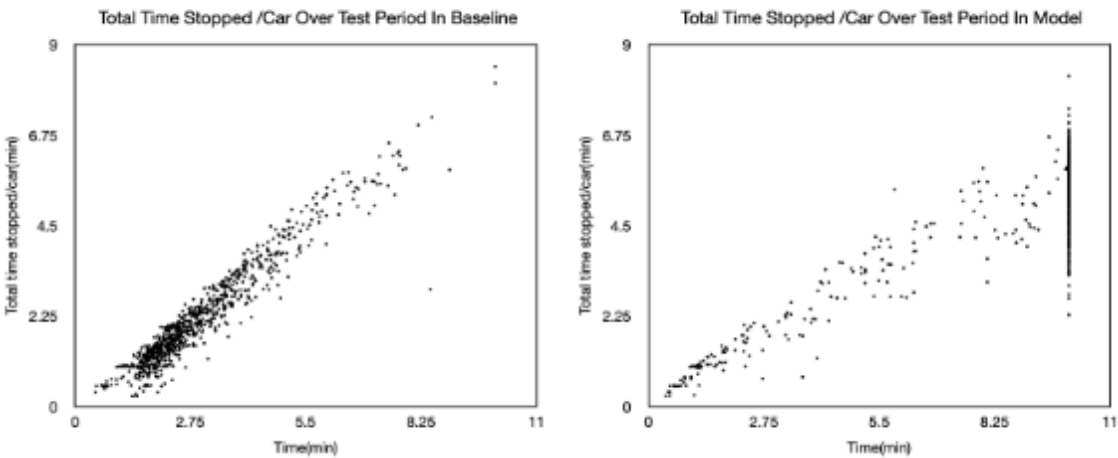

Fig 8: Total time stopped per car in baseline (left) vs mode (right). In the baseline, the majority of vehicles undergo serious collisions earlier on in the testing period, eliminating them from the environment eventually. The model prevents collisions; the majority of vehicles are preserved at the end of the testing period; they undergo more stops as there are more vehicles (more congestion) remaining towards the end of the testing environment.

The model also yields corresponding environmental benefits, with a 39.33% improvement in fuel efficiency per vehicle and an 88% reduction in carbon emissions per vehicle (Figs. 9-10). All of the p values were $10^{-23}$ or less except for time stopped which had a p value of $10^{-5}$. Performance improvements were validated, even without additional training, in a large sized urban simulation environment with 67 traffic intersections and 1896 vehicles (Supplement Figs. 5-11). However, fine tuning to each environment with it's particular road network and intersection type and number is highly recommended for the best performance.

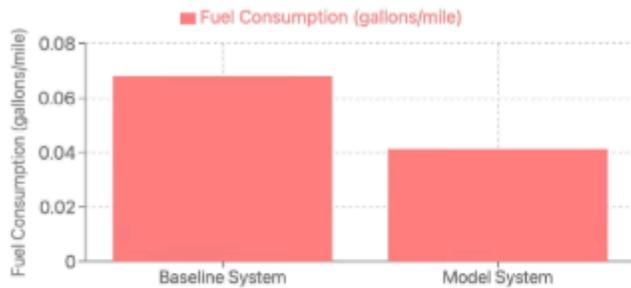

Fig 9: Vehicle Fuel consumption(gallons/mile), baseline vs model. There was a 39.33% improvement in fuel efficiency per vehicle with the model

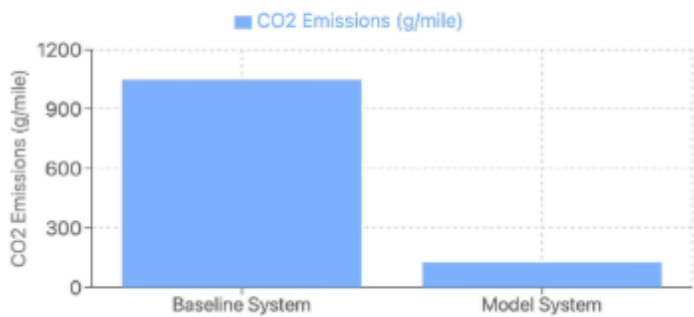

Fig 10. Vehicle CO2 emissions (g/mile) in baseline vs model. The model improved CO2 emissions by 88%

**DISCUSSION**

This 3D traffic simulation allows for intuitive game-like visualization, providing improved interpretability, accessibility and explainability for technical and nontechnical personnel.  The random generation of urban/city area specific road network and intersections, allows for  prompt experimentation, particularly with the built in RL models in Unity. Performance  improvements by the RL models including PPO were predominantly achieved through optimized  traffic light timing rather than adjusted speed limits. The latter was likely constrained by the  proximity of intersections in the downtown simulation environment, which in combination with  high traffic volumes, inherently limited travel speeds. Highways were not included in the current  study, similar to many other urban macro simulation environment studies, due to lack of ability  (via traffic lights or other ways) to sufficiently control traffic. However highway ramp metering  can be modeled by utilizing many of Unity's road and highway assets.

 It is possible to incorporate local real-world environments into the simulation environment using available infrastructure and sensor data. Road network data can be obtained from various mapping services (e.g., Google Maps, OpenStreetMap, MapBox, TomTom, and HERE). Traffic information can be collected via cameras, radars, or GPS. Connected vehicle data are available through services like Otonomo, and real-time weather information can be sourced from OpenWeather or Tomorrow.ai. Moreover, road condition and collision data can be obtained from Waze.

The custom PPO model in this study achieved over 300% improvement in average distance traveled. This differs from other studies showing at most a 15% improvement[38]. A significant portion of this discrepancy is due to the fact that there were more vehicles in the model optimized environment, secondary to reduced collisions from the safety-priority model. However the average time stopped increase was disproportional at only a third of the increase in average distance travelled (110% vs 345%), demonstrating that even within the surviving vehicles, traffic flow was optimized to reduce congestion. This was also borne out by the 88% reduction in carbon emissions per vehicle in the optimized environment, despite travelling longer distances. Another difference lay in some studies[38] choosing the phase (e.g. green for east west flows with red for north south and vice versa but no change in 30 sec green time) rather than green light time to modify. The latter was chosen in this study in order to avoid frequent transition phases (yellow) if the same phase needed to be chosen over and over again, and also resulted in more granular control.

While the elevated collision frequency employed in the current study exceeds real-world rates, it enables statistically significant results to be obtained for what would otherwise be rare events. The vehicle dynamics can be modified to conservatively minimize collisions to real life rates, so as to prioritize traffic congestion while still evaluating the effect on collisions. Balancing congestion reduction and collision prevention required extensive reward shaping, as improvements in one metric often degraded the other; slower moving traffic was safer but more congested. The eventual custom rewards played a heavy role in congestion improvement.

The biggest advantage of this game engine-based simulation environment is the ability to model realistic physics enabled traffic flow as well as collisions that are not prevalent in current simulation environments, and tend to be more accurate than coded collisions. This does result in higher computational complexity compared to current 2D environments but is positioned to benefit from lowering costs of storage and processing power. Attempts at implementing curriculum learning with gradual increases in green light durations and vehicle density were constrained by prohibitively long training periods. Increased computational overhead caused by processing high dimensional observations from the environment every frame (supplement section 5) were addressed by random 50 point sampling for each frame and the use of the Unity no-graphics builder.

Future refinements could include mitigating computational constraints through behavior cloning, model pre-training with diverse datasets, and network pruning techniques that selectively remove redundant weights, neurons, or filters to reduce the overall parameter count in the RL architecture. Overfitting to any one simulation city could be mitigated by encoding random environmental changes, such as vehicle respawning or orientation variations. Expanded simulations could incorporate pedestrian walkways, bicycle lanes, and construction areas.

**CONCLUSION**

A custom-reward PPO model in a sample 3D environment demonstrated marked collision reduction as well as substantially improved traffic congestion despite prioritizing safety over throughput, in keeping with the vision for zero fatality rates espoused by DOT. Future traffic related work could focus on transitioning to real-world datasets for offline training and subsequent translations to limited pilot implementations. Creation of physics informed simulations through game engines are intuitive due to graphic visualization feedback and analysis. Apart from congestion relief, conflict evolution and resolution could be studied not only in traffic, but other network problems including energy grids or communication relay infrastructure.

*Data Availability*

Code for the creation of simulation data and customization of models, as well as testing datasets are available in the following GitHub repository: https://github.com/lgh1000/Traf.

*Author Contributions*

M.N. conceived the study, developed the simulation environment, designed and implemented the reinforcement learning models, analyzed the results, and wrote the manuscript.

*Competing Interests*

The author declares no competing interests.

Actuated_Signal_Controller_For_Heterogeneous_Traffic_Having_Limited_Lane_Discipline

# SUPPLEMENT

Video: Simulated 3D city-wide environment comparison with 2D simulation environments:
https://www.youtube.com/watch?v=GT-ogvCqASA

## SECTION 1 - ENVIRONMENT

**City Generation:** The urban environment was generated using the city generator asset framework. The `distCenter` parameter (ranging from 150 to 350 units) defined the urban density characteristics, with smaller values creating downtown areas and larger values creating suburban regions. Each city was enclosed by a border mesh that defined the outer limits of the navigable area.

Building placement utilized a typology-based categorization system with nine distinct building types based on their urban function (residential, commercial, corner, block-spanning). Building selection incorporated probability-based residential zone distribution, with higher residential density occurring at distances greater than 400 units from the city center and when the random allocation percentage was below 30%. Four distinct city scales were as follows:

1. Very Small: A minimal urban area with a single central block at (0,0,0) and a distance center parameter of 150 units.
2. Small: A moderate sized compact environment with a distance center of 200 units, featuring two large blocks positioned either at [(0,0,0), (0,0,300)] or [(-150,0,150), (150,0,150)] based on random selection.
3. Medium: A mid-sized environment with a distance center of 300 units, containing four large blocks with positions determined by one of four different layout patterns selected randomly.
4. Large: An extensive environment with a distance center of 350 units, containing six large blocks with positions determined by one of four layout configurations.

**Building Generation System:** Buildings were categorized into nine distinct groups based on their position and function in the urban landscape:

- BB: Street buildings in suburban areas
- BC: Downtown buildings
- BR: Residential buildings in suburban areas
- DC: Corner buildings spanning both sides of a block
- EB: Corner buildings in suburban areas
- EC: Downtown corner buildings
- MB: Buildings occupying both sides of a block
- BK: Buildings occupying an entire block
- SB: Large buildings occupying larger blocks

Buildings were positioned according to these categories, using a combination of offset calculations, raycasting for surface detection, and transform operations to ensure proper alignment with the terrain and roads. Dimensional information from mesh bounds was extracted to inform placement decisions. Building scale was adjusted dynamically to fit available spaces.

**Mesh Optimization:** For performance optimization, multiple meshes are combined into unified objects. This utility performed vertex welding, normal recalculation, and UV mapping to reduce draw calls. A vertex limit of 65,000 was enforced to comply with Unity's mesh limitations.

**Visual and Physical Wheel Synchronization:** To maintain visual fidelity, wheel transforms were synchronized with the physical wheel colliders on each update. This synchronization ensured that the visible wheels accurately represented the physical simulation state, including rotation, suspension compression, and steering angle.

Vehicle parameters and metrics:

- Vehicle mass: 4000 units
- Suspension spring strength: 25,000 (range: 10,000-60,000)
- Suspension damping: 1,500 (range: 1,000-6,000)
- Engine power: 120 (range: 60-200)
- Braking power: 8 (range: 5-10)
- Speed limit: 30 (range: 20-35)
- Maximum steering angle: 35-72° (dynamically calculated based on wheelbase length)

The center of mass was positioned at (0.0, -0.05, 0.0) relative to the vehicle's origin to improve stability and prevent roll-over during cornering.

Vehicle speed distributions were tracked using a binning system with 5-unit intervals from 0-35 units/sec for an urban environment:

- Speed0-5: Very slow or stopped traffic
- Speed5-10: Crawling traffic
- Speed10-15: Slow traffic
- Speed15-20: Moderate traffic
- Speed20-25: Normal traffic
- Speed25-30: Free-Flowing traffic
- Speed30-35: Fast traffic

Time spent in each speed range was continuously recorded, providing detailed traffic flow characteristics for post-simulation analysis.

**Gizmo based visualization in Unity:** Visualization provided several key benefits during environment development:

1. Spatial Orientation: Each waypoint was represented by a solid green sphere (1 unit radius) surrounded by a wire sphere (2 unit radius), making them easily identifiable in the scene view.
2. Path Connectivity: Green lines connected sequential waypoints, visualizing the complete vehicle path and highlighting any gaps or discontinuities.
3. Debugging Navigation: The visualization allowed developers to visually trace vehicle paths through the environment, identify problematic intersections, and verify proper waypoint positioning and orientation.
4. Intersection Validation: By observing the terminal waypoints at intersections, developers could verify that proper connectivity was established between different path segments.
5. Editor-Only Overhead: Since Gizmos only render in the Unity Editor, they added no performance overhead to the runtime simulation.

The Gizmo system was vital during the development process, as it made the otherwise invisible navigation infrastructure visually apparent, allowing for efficient environment construction and debugging without affecting runtime performance.

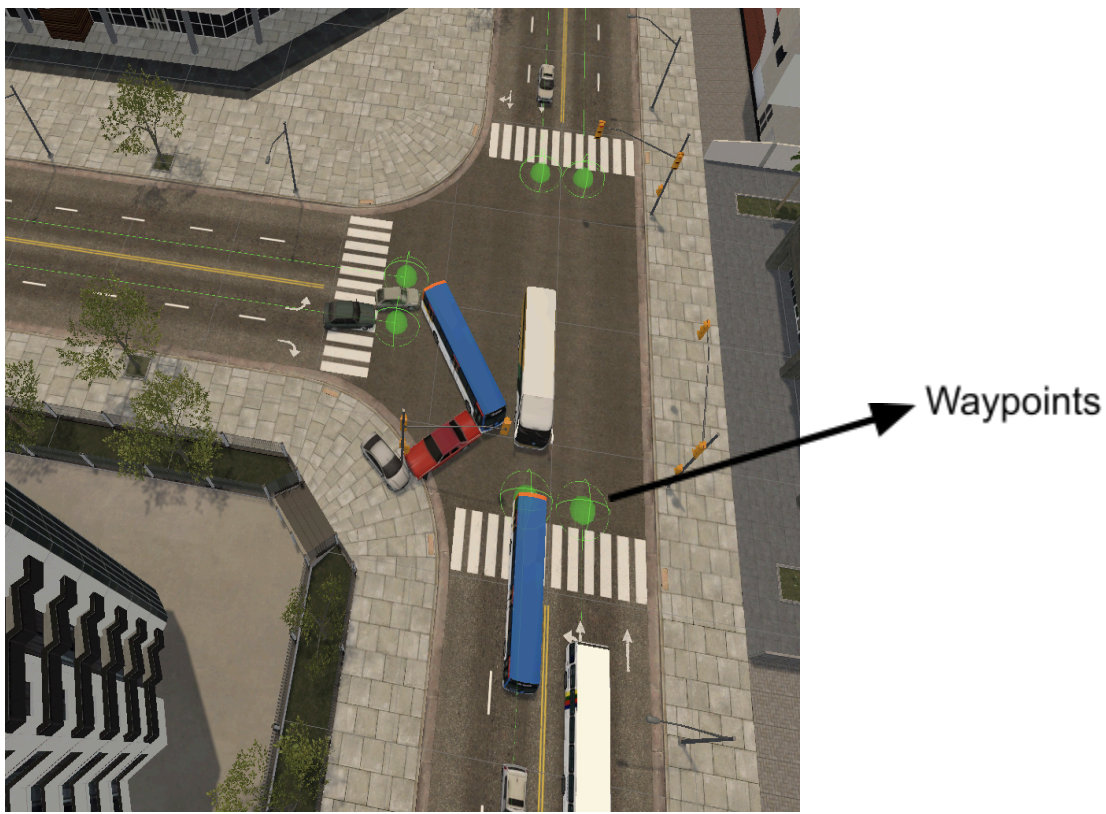

Supplement Fig 1 Severe major collision involving multiple vehicles.

# SECTION 2 - MODEL

Reinforcement learning model schema

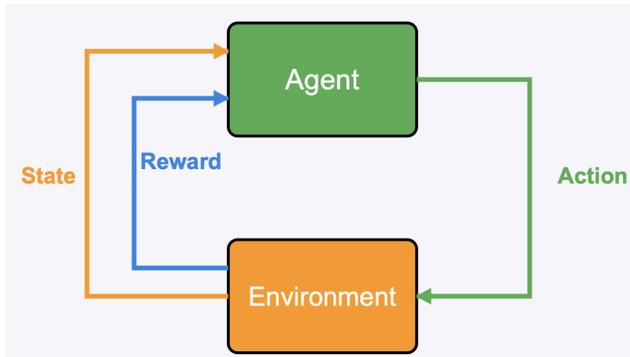

Supplement Fig 2. Reinforcement learning model. The agent is the decision maker. For each action made by the agent on the environment, the environment provides a reward. The state of the agent in the environment is updated based on this feedback. Using the new learnt policy (strategy), the agent takes a new action.

# SECTION 3 - RESULTS

## Performance of PPO (Proximal Policy Optimization, SAC (Soft Actor Critic) and MA_POCA (Multi-Agent POsthumous Credit Assignment)

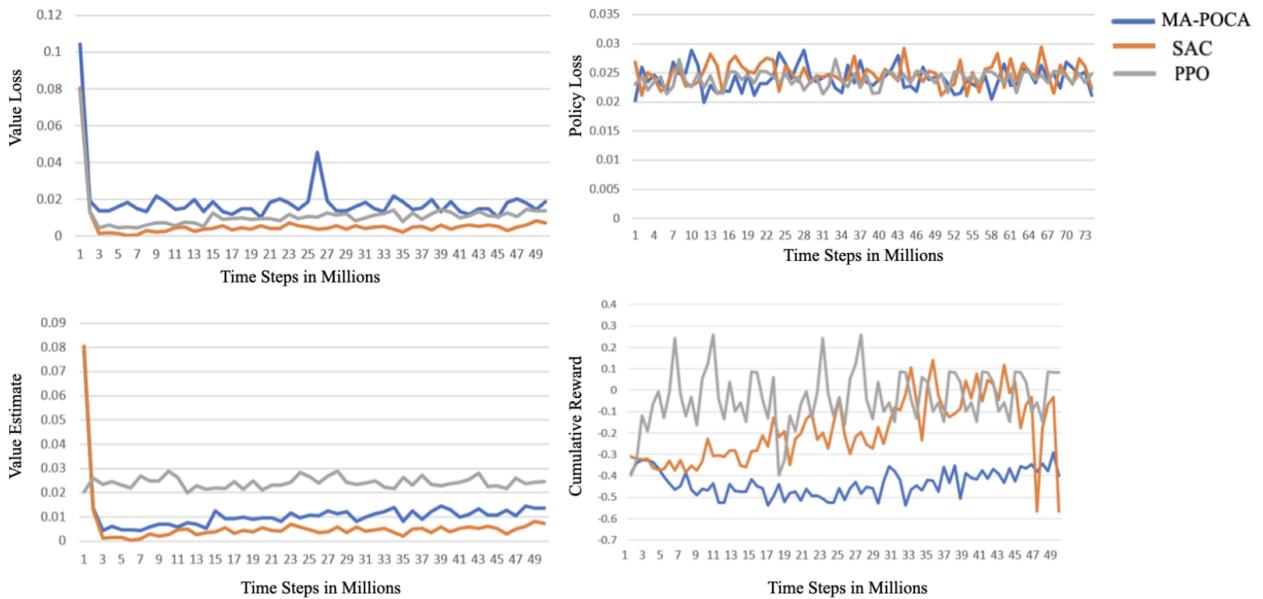

Supplement Fig 3: Value Loss, Policy Loss, Value Estimate and Cumulative reward comparison in PPO, SAC and MA-POCA (536 vehicles). PPO demonstrated earlier, higher, and more consistent cumulative rewards with higher value estimates and moderate value loss compared to SAC and MA-PPO

## PPO Training Metrics

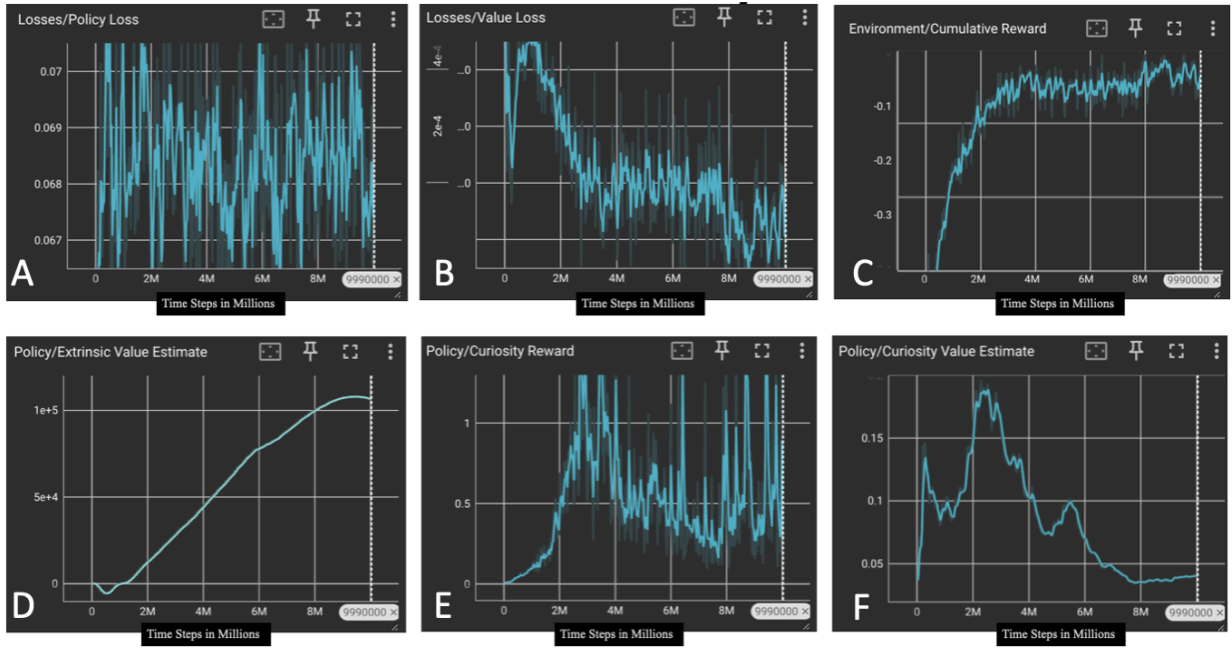

Supplement Fig 4: Training Metrics of parameters for PPO in 897 vehicles. A. Policy loss: how much the policy (process for deciding actions) is changing. Tends to be below 1 if there is stable training • B. Value loss: how well the model is able to predict the value of each state. It tends to increase with reward and decrease when rewards stabilize. C. Cumulative reward: Accumulated rewards increases till plateau. D. Value Estimate: how much future reward the agent predicts itself receiving. If it increases linearly, it is indicative of good policy and learning. E. Curiosity reward for exploration enables early exploration and later decreases when transitioning to exploitation. F. Curiosity Value Estimate: prediction of curiosity reward. It follows curiosity reward.

## SECTION 4 - RESULTS VALIDATION

Results on a different environment with 1896 vehicles and 67 traffic lights:

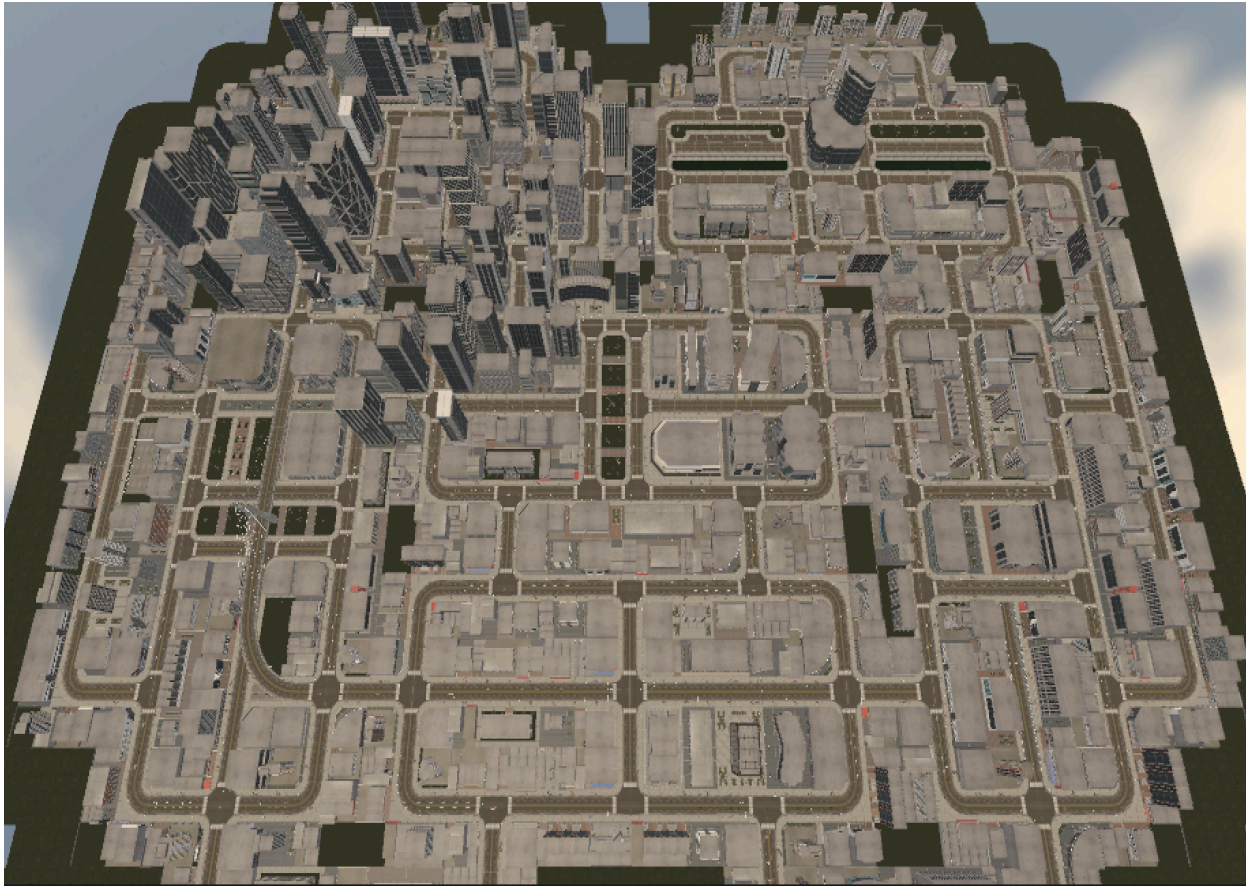

Supplement Fig 5: 67 traffic light, 1896 vehicles environment, with downtown section.

Each environment configuration necessitates its own dedicated training regimen, or at minimum, fine-tuning through transfer learning to achieve optimal performance. To evaluate model robustness and demonstrate cross-validity, the model trained on the primary environment (46 traffic lights, 879 vehicles) was tested in a more complex validation environment (67 traffic lights, 1896 vehicles) without additional training. This cross-environment evaluation yielded comparable patterns of improvement, albeit with reduced magnitude in distance metrics (345% improvement in the original environment versus 21% improvement in the validation environment, p=0.0073, Fig. 7 main paper, Supplement Fig. 11). This performance differential is consistent with expectations given the substantially increased traffic density (>100% increase in vehicle count); moreover, the reward function was calibrated to prioritize collision avoidance over congestion mitigation in the primary environment.

Interestingly, peak performance occurred at different speed thresholds across environments: the 20-25 mph range exhibited the greatest improvement in the validation environment, compared to 25-30 mph in the original environment. This discrepancy likely reflects the increased congestion resulting from the more than doubled vehicle density in the validation scenario (Fig. 7 main paper, Supplement Fig. 10).

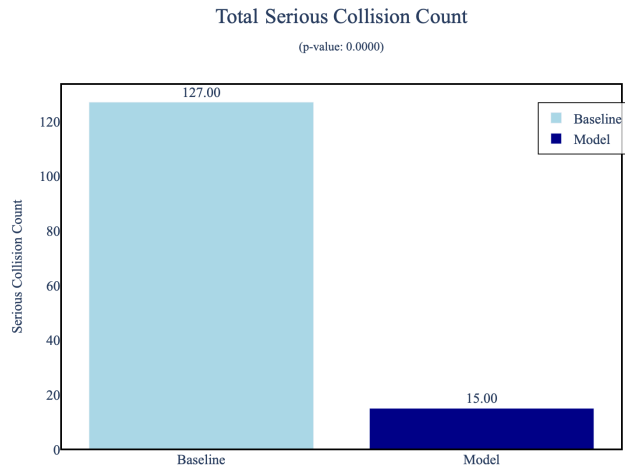
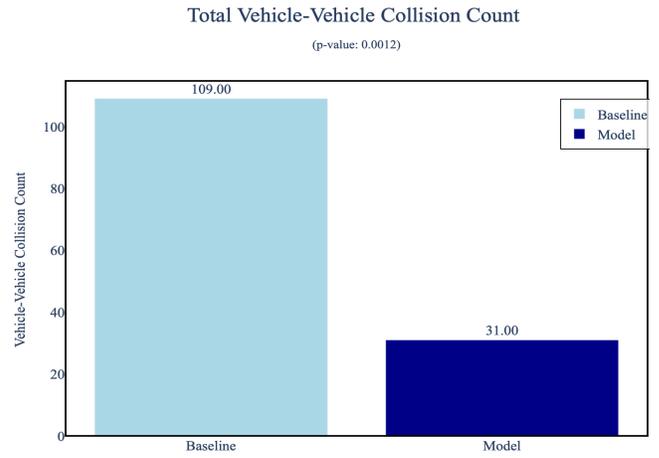

Supplement Fig 6: Total serious collision count in baseline vs model. Model improved the serious collisions from 127 to 15, an 88% improvement; p 0.0000

Supplement Fig 7: Total non-serious vehicle to vehicle collision count in baseline vs model. Improvement via model was 109 to 31, a 71% improvement, p 0.0012.

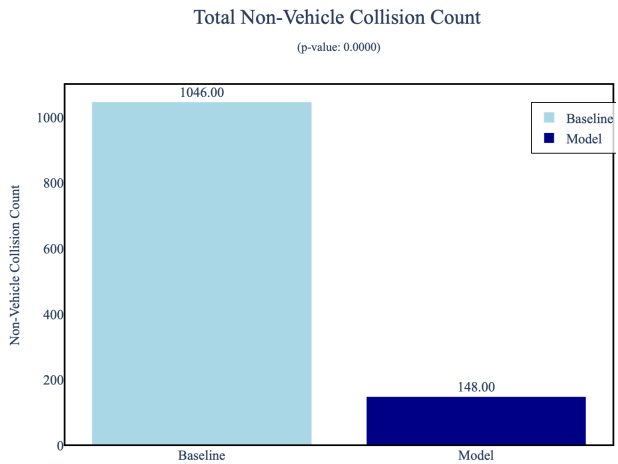
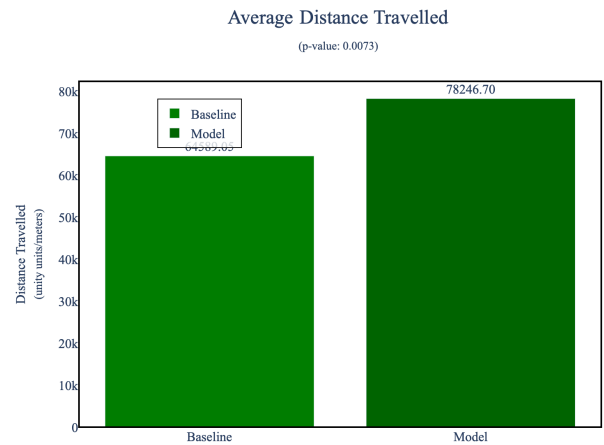

Supplement Fig 8: Minor Vehicle to non-vehicle collision count in baseline vs model. Improvement via model was 1046 to 148, an 85.8% improvement, p 0.0012.

Supplement Fig 9: Average distance travelled in baseline vs model. Improvement via model was 78,246.70 meters/unity units to 64589.05m, a 21% improvement, p 0.0073

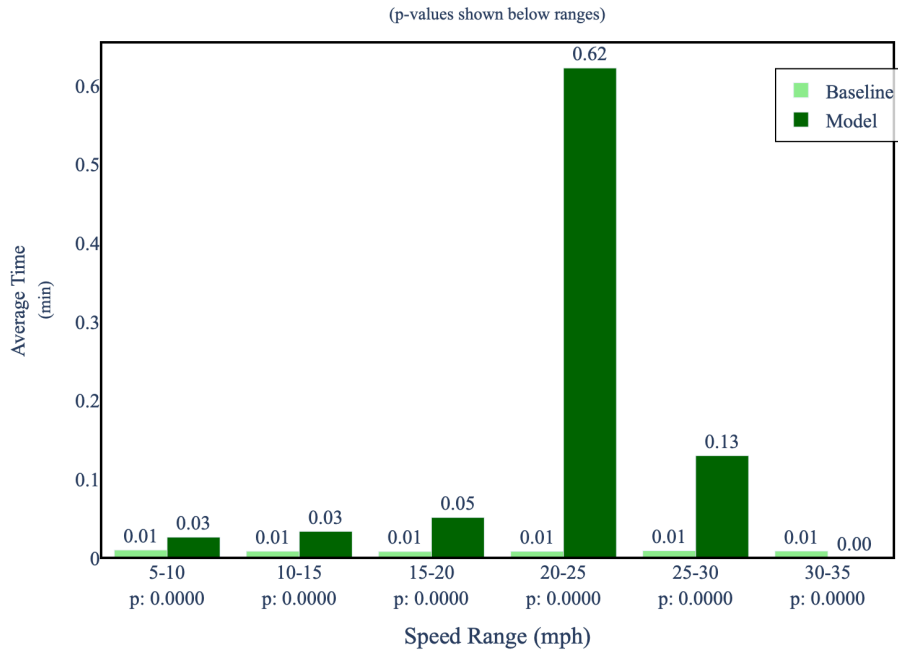

Supplement Fig 10: Average time spent (min) in speed ranges 0-5mph, 5-10mph, 10-15mph, 15-20mph, 20-25mph, 25-30mph, 30-35mph in baseline vs model. Majority of the increase was in 20-25mph, a 6100% improvement in this speed range, with all p values <0.00001

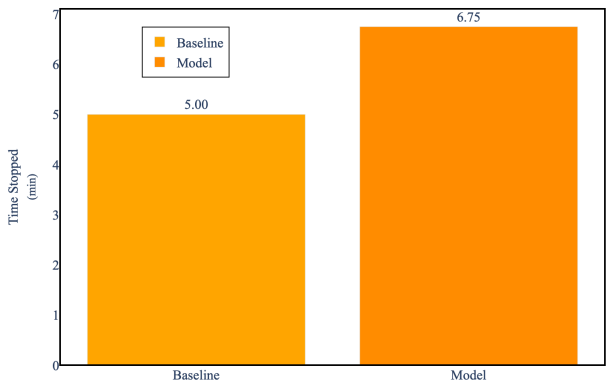

Supplement Fig 11: Average time stopped(min) in baseline vs model. Average time stopped increased from 5 minutes to 6.75 minutes, a 35% increase with p <0.00001. Similar to the original environment, the majority of vehicles in the baseline testing environment undergo serious collisions earlier on in the testing period, eliminating them from the environment eventually. The model prevents collisions; the majority of vehicles are preserved at the end of the testing period; they undergo more stops as there are more vehicles (more congestion) remaining towards the end of the testing environment.

# SECTION 5 - OTHER EXPERIMENTS

Curriculum Learning Proximal Policy Optimization (CL-PPO) was implemented with progressively increasing green light durations and vehicle density, although this approach was constrained by significantly extended training periods and the necessity to simultaneously optimize multiple parameters including traffic light cycles, speed restrictions, and vehicle quantities (Supplement Fig. 12-14). Weights obtained by early CL-PPO experiments were incorporated into the final PPO model.

A fundamental trade-off emerged between congestion mitigation and collision prevention, with improvements in one metric typically resulting in deterioration of the other (Supplement Fig. 12,13).

Inconsistent frame rates (ranging from 0.00059 to 0.003 seconds) compromised the reliability of temporal measurements and metrics calculations, consequently affecting reward computation. This challenge was addressed by anchoring rewards and metrics to real-time calculations rather than simulation time (which inconsistently operated at 1-20x real-time speed, depending on observational load), albeit at the cost of increased training duration.

The observational space presented significant computational challenges, as even essential observations (879 vehicles × 2 parameters + 46 traffic lights × 4 parameters) totaled 1,942 inputs processed approximately every 0.0005 seconds. This computational burden was mitigated through temporal distribution of observations and frame stacking. When these approaches proved insufficient, a sampling strategy was implemented whereby only 50 randomly selected vehicles per time frame were observed. While effective, this solution remained computationally intensive, necessitating the implementation of a Unity Build without graphics to accelerate processing.

More complex reward structures yielded diminished performance outcomes, whereas a bias toward high-frequency rewards demonstrated greater effectiveness.

The dual problems of excessive collision frequency and infrequent collision detection were resolved through systematic experimentation, with the optimal configuration consisting of three raycasts (one forward and two at 45° angles).

Several methodological challenges were encountered, including initial convergence to a local minimum where traffic light durations were constrained to 0-2 seconds to avoid the substantial penalties associated with serious collisions (-3 per serious collision, which inhibited exploration). These challenges were addressed through: 1) reward function recalibration to establish appropriate balance between collision penalties and distance-traveled incentives, and 2) enhanced exploration facilitated by incorporating curiosity-driven rewards and increased entropy coefficients (beta)

A low collision environment was unable to establish statistical significance due to low sample size (Supplement Fig 15). As such a high collision environment was created to evaluate for statistical significance (main paper)

SERIOUS COLLISIONS WITH TRAFFIC GREEN LIGHT DURATION
IN CURRICULUM LEARNING PPO (879 vehicles)

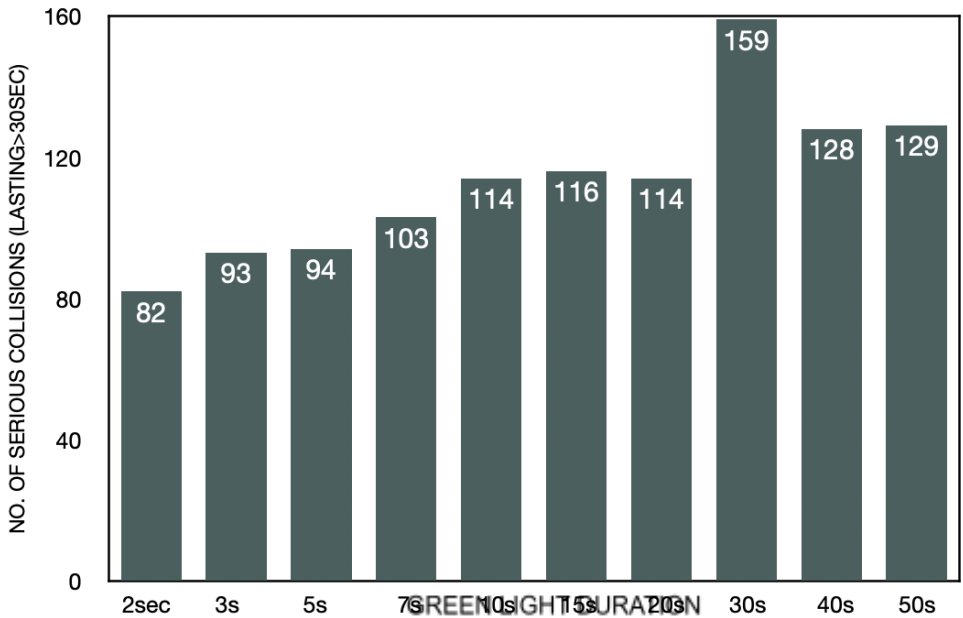

Supplement Fig 12. Number of serious collisions with increasing green traffic light duration in curriculum learning (CL) PPO. There was a trend towards increasing collisions with increasing green light

TOTAL TIME STOPPED WITH TRAFFIC GREEN LIGHT DURATION
IN CURRICULUM LEARNING PPO (879 vehicles)

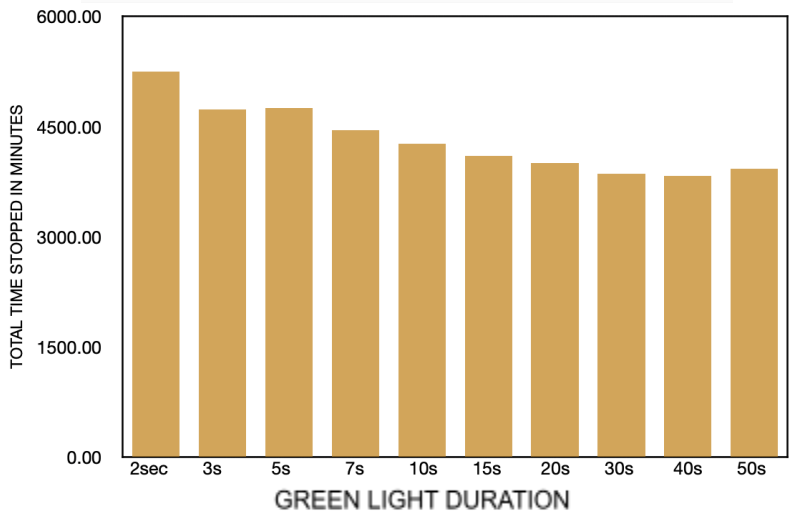

Supplement Fig 13. Total time stopped in seconds with increasing green light durations in 879 vehicles utilizing curriculum learning PPO. There was a trend towards decreasing time stopped with increasing green light duration till 30-40 sec after which it started increasing again.

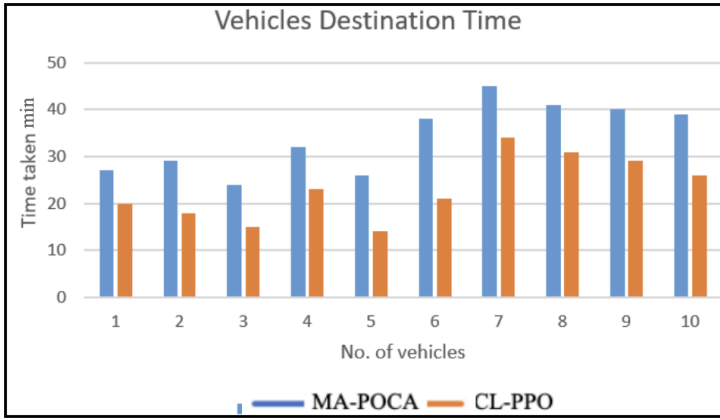

Supplement Fig 14. Time taken in minutes of increasing number of vehicles in curriculum Learning (CL) PPO vs MA-POCA. CL-PPO performed better with less time taken

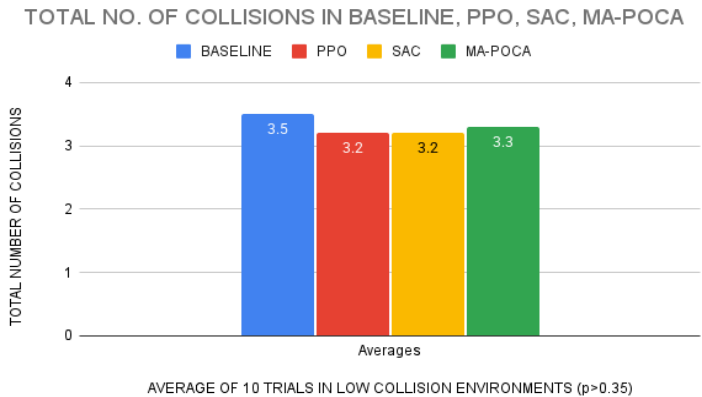

Supplement Fig 15. Number of total collisions in a more realistic scenario. Baseline was programmed with low tendency to collide amongst 536 vehicles. Testing with PPO, SAC and MA-POCA revealed no significant difference (p>=0.35) given the small sample sizes.

## SECTION 6 - HYPERPARAMETERS AND PACKAGES

**HYPERPARAMETERS:**
default_settings: null
behaviors:
　TrafficOptimization_925A_284x1Ob_2_exp:
　　trainer_type: ppo
　　hyperparameters:
　　　batch_size: 2560
　　　buffer_size: 102400
　　　learning_rate: 0.0003
　　　beta: 0.05
　　　epsilon: 0.2
　　　lambd: 0.95
　　　num_epoch: 3
　　　shared_critic: false
　　　learning_rate_schedule: linear
　　　beta_schedule: linear

```yaml
      epsilon_schedule: linear
    network_settings:
      normalize: true
      hidden_units: 512
      num_layers: 2
      vis_encode_type: simple
      memory: null
      goal_conditioning_type: hyper
      deterministic: false
    reward_signals:
      extrinsic:
        gamma: 0.99
        strength: 1.0
        network_settings:
          normalize: true
          hidden_units: 128
          num_layers: 2
          vis_encode_type: simple
          memory: null
          goal_conditioning_type: hyper
          deterministic: false
      curiosity:
        gamma: 0.99
        strength: 0.02
        network_settings:
          normalize: true
          hidden_units: 256
          num_layers: 2
          vis_encode_type: simple
          memory: null
          goal_conditioning_type: hyper
          deterministic: false
        learning_rate: 0.0003
        encoding_size: null
    init_path: null
    keep_checkpoints: 5
    checkpoint_interval: 500000
    max_steps: 10000000
    time_horizon: 512
    summary_freq: 30000
    threaded: false
    self_play: null
    behavioral_cloning: null
env_settings:
  env_path: null
  env_args: null
  base_port: 5005
  num_envs: 1
  num_areas: 1
  seed: -1
  max_lifetime_restarts: 10
  restarts_rate_limit_n: 1
  restarts_rate_limit_period_s: 60
engine_settings:
  width: 84
  height: 84
  quality_level: 5
```

```
    time_scale: 20.0
    target_frame_rate: -1
    capture_frame_rate: 60
    no_graphics: true
environment_parameters: null
checkpoint_settings:
    run_id: TrafficOptimization_925A_284x1Ob_2_exp
    initialize_from: null
    load_model: false
    resume: false
    force: false
    train_model: false
    inference: false
    results_dir: results
torch_settings:
    device: null
debug: true
```

## PACKAGES

```
#
# Name                    Version           Build  Channel
absl-py                   2.1.0             pypi_0    pypi
astunparse                1.6.3             pypi_0    pypi
attrs                     23.2.0            pypi_0    pypi
ca-certificates           2023.12.12        hca03da5_0
cachetools                5.3.3             pypi_0    pypi
cattrs                    1.5.0             pypi_0    pypi
certifi                   2024.2.2          pypi_0    pypi
charset-normalizer        3.3.2             pypi_0    pypi
cloudpickle               3.0.0             pypi_0    pypi
dm-tree                   0.1.8             pypi_0    pypi
expat                     2.5.0             h313beb8_0
filelock                  3.13.1            pypi_0    pypi
flatbuffers               24.3.7            pypi_0    pypi
fsspec                    2024.3.1          pypi_0    pypi
gast                      0.4.0             pypi_0    pypi
google-auth               2.29.0            pypi_0    pypi
google-auth-oauthlib      1.0.0             pypi_0    pypi
google-pasta              0.2.0             pypi_0    pypi
grpcio                    1.62.1            pypi_0    pypi
gym                       0.26.2            pypi_0    pypi
gym-notices               0.0.8             pypi_0    pypi
h5py                      3.10.0            pypi_0    pypi
idna                      3.6               pypi_0    pypi
importlib-metadata        7.0.2             pypi_0    pypi
jinja2                    3.1.3             pypi_0    pypi
keras                     3.2.1             pypi_0    pypi
libclang                  18.1.1            pypi_0    pypi
libcxx                    14.0.6            h848a8c0_0
libffi                    3.4.4             hca03da5_0
libiconv                  1.16              h1a28f6b_2
markdown                  3.6               pypi_0    pypi
markdown-it-py            3.0.0             pypi_0    pypi
markupsafe                2.1.5             pypi_0    pypi
mdurl                     0.1.2             pypi_0    pypi
ml-dtypes                 0.3.2             pypi_0    pypi
```

```
mlagents                 0.30.0                    pypi_0    pypi
mlagents-envs            0.30.0                    pypi_0    pypi
mpmath                   1.3.0                     pypi_0    pypi
namex                    0.0.7                     pypi_0    pypi
ncurses                  6.4                  h313beb8_0
networkx                 3.2.1                     pypi_0    pypi
numpy                    1.23.3                    pypi_0    pypi
oauthlib                 3.2.2                     pypi_0    pypi
onnx                     1.16.0                    pypi_0    pypi
openssl                  1.1.1w               h1a28f6b_0
opt-einsum               3.3.0                     pypi_0    pypi
optree                   0.11.0                    pypi_0    pypi
packaging                24.0                      pypi_0    pypi
pettingzoo               1.15.0                    pypi_0    pypi
pillow                   10.2.0                    pypi_0    pypi
pip                      23.3.1           py39hca03da5_0
protobuf                 3.20.3                    pypi_0    pypi
pyasn1                   0.6.0                     pypi_0    pypi
pyasn1-modules           0.4.0                     pypi_0    pypi
pygments                 2.17.2                    pypi_0    pypi
python                   3.9.7                hc70090a_1
pyyaml                   6.0.1                     pypi_0    pypi
readline                 8.2                  h1a28f6b_0
requests                 2.31.0                    pypi_0    pypi
requests-oauthlib        2.0.0                     pypi_0    pypi
rich                     13.7.1                    pypi_0    pypi
rsa                      4.9                       pypi_0    pypi
setuptools               68.2.2           py39hca03da5_0
six                      1.16.0                    pypi_0    pypi
sqlite                   3.41.2               h80987f9_0
sympy                    1.12                      pypi_0    pypi
tensorboard              2.16.2                    pypi_0    pypi
tensorboard-data-server  0.7.2                     pypi_0    pypi
tensorflow               2.16.1                    pypi_0    pypi
tensorflow-estimator     2.13.0                    pypi_0    pypi
tensorflow-io-gcs-filesystem 0.36.0                pypi_0    pypi
tensorflow-macos         2.13.0                    pypi_0    pypi
termcolor                2.4.0                     pypi_0    pypi
tk                       8.6.12               hb8d0fd4_0
torch                    1.11.0                    pypi_0    pypi
typing-extensions        4.5.0                     pypi_0    pypi
tzdata                   2024a                h04d1e81_0
urllib3                  2.2.1                     pypi_0    pypi
werkzeug                 3.0.1                     pypi_0    pypi
wheel                    0.41.2           py39hca03da5_0
wrapt                    1.14.1                    pypi_0    pypi
xz                       5.4.6                h80987f9_0
zipp                     3.18.1                    pypi_0    pypi
zlib                     1.2.13               h5a0b063_0
```